\pdfoutput=1
\documentclass[10pt,twocolumn,letterpaper]{article}

\usepackage{cvpr}
\usepackage{times}
\usepackage{epsfig}
\usepackage{graphicx}
\usepackage{amsmath}
\usepackage{amssymb}

\usepackage[numbers,sort,compress]{natbib}
\usepackage{bm}
\usepackage{subcaption}
\usepackage{multirow}
\usepackage{diagbox}

\newcommand{\TwoColTwoSubfigWidth}{0.38\columnwidth}

\newcommand{\FigWidth}{0.8\columnwidth}

\newcommand{\KL}{D_{\text{KL}}}

\newcommand{\cX}{\mathcal{X}}
\newcommand{\cY}{\mathcal{Y}}
\newcommand{\cD}{\mathcal{D}}

\newcommand{\defeq}{\overset{\text{def}}{=}}

\usepackage[pagebackref=true,breaklinks=true,colorlinks,bookmarks=false]{hyperref}

\usepackage[capitalize,noabbrev]{cleveref}

\cvprfinalcopy 

\def\cvprPaperID{6783} 

\ifcvprfinal\fi
\begin{document}

\title{Few-Shot Self Reminder to Overcome Catastrophic Forgetting}


\author{
Junfeng Wen\textsuperscript{$\dag,\ddag$,}\thanks{Work done while interning at Borealis AI}\qquad
Yanshuai Cao\textsuperscript{$\dag$}\qquad
Ruitong Huang\textsuperscript{$\dag$}\\
\textsuperscript{$\dag$}Borealis AI\qquad
\textsuperscript{$\ddag$}University of Alberta\\
  {
  \tt\small junfengwen@gmail.com\qquad
  \{yanshuai.cao,ruitong.huang\}@borealisai.com
  }
}

\maketitle

\begin{abstract}
Deep neural networks are known to suffer the catastrophic forgetting problem, where they tend to forget the knowledge from the previous tasks when sequentially learning new tasks. 
Such failure hinders the application of deep learning based vision system in continual learning settings.
In this work, we present a simple yet surprisingly effective way of preventing catastrophic forgetting. 
Our method, called Few-shot Self Reminder~(FSR),
regularizes the neural net from changing its learned behaviour by performing logit matching on selected samples kept in episodic memory from the old tasks.
Surprisingly, this simplistic approach only requires to retrain a small amount of data in order to outperform previous methods in knowledge retention. 
We demonstrate the superiority of our method to the previous ones in two different continual learning settings on popular benchmarks, as well as a new continual learning problem where tasks are designed to be more dissimilar. 

\end{abstract}

\section{Introduction}
Both human vision system, as well as many real-world applications of machine vision, need to continually adapt as the environment changes or as new visual concepts are required. 
The change in the inputs or outputs of the vision system presents new tasks to be learned. Deep learning based vision system has more difficulties than human vision in dealing with this sequential learning situation because of the so-called catastrophic forgetting problem.
In sequential learning of multiple tasks, artificial neural networks catastrophically forget previous knowledge when new tasks are learned~\cite{mccloskey1989catastrophic,mcclelland1995there,french1999catastrophic}, as the optimization in a later stage could adapt the shared parameters and representations in ways that harm the old tasks.
This failure hampers the application of deep models since it indicates that they are incapable of maintaining knowledge when facing new environments or new tasks in the same environment. 

Many different ways to address this problem have been explored in the literature.
One straightforward approach would be to jointly train on old and new tasks as in multi-task learning~\cite{Caruana1997}. 
However, storing all the previous data could be resource demanding in practice. 
Hence alternative methods that store models rather than historical data have been proposed in the literature~\cite{kirkpatrick2017overcoming,zenke2017continual,lopez2017gradient,lee2017overcoming,yoon2018lifelong,he2018overcoming}. 
For example, Elastic Weight Consolidation~(EWC)~\cite{kirkpatrick2017overcoming} stores the previous model to estimate the sensitivity of the previous task loss to different parameters, and penalizes the model parameter changes from one task to the next according to the different sensitivities. 
\citet{shin2017continual} replaces the storage of the previous data by training GANs to generate fake historical data.

Besides the complications of these methods, which usually leads to an exhausted hyperparameter tuning, such approaches also suffer a common deficiency: in order to cover the historical data reasonably well, the size of the extra stored models is usually a significant cost,
given that current state-of-the-art neural networks usually involve millions of parameters.
Thus, such methods may not necessarily save its storage cost, especially when saving a small subset of ``anchor'' points from the historical data is sufficient for this problem, as we will show later. 

In this work, we propose Few-shot Self Reminder~(FSR). 
FSR is frustratingly simple but surprisingly effective in practice, demonstrating that  
it is possible to address catastrophic forgetting by storing and reusing very few previous data without incurring significant memory cost. 
It also clarifies a potential misconception that storing old data is too costly for joint training hence we need more complex approaches. 
The idea of FSR is to place the regularization on the function mapping instead of its parameters. 
It does so with a small episodic memory of previous data and their corresponding output logits by the model.
This idea is adopted from the model compression community~\cite{bucila2006model,ba2014deep,hinton2015distilling}, but used in a very different manner: instead of distillation/logit matching on a large amount of input-output data, we apply them on a small subset to maintain its desired behaviour on a much larger dataset.

To summarize our key empirical findings:
\begin{enumerate}
\item On continual input distribution change settings, FSR outperforms previous methods such as LwF~\cite{li2017learning}, iCaRL~\cite{rebuffi2017icarl}, EWC~\cite{kirkpatrick2017overcoming}, and Generative Replay~\cite{shin2017continual}. 
Specifically, to achieve a similar accuracy to EWC for 5 Permuted MNIST tasks, FSR needs as little as $5\%$ of memory requirement of EWC.
\item FSR only needs to memorize as little as one data point per class to reduce the forgetting to be slow and gradual rather than ``catastrophic''.
\item In class incremental settings, FSR can be comparable with or superior to HAT~\cite{serra2018overcoming}, when tasks share similarities for knowledge to transfer via feature re-use.
\end{enumerate}

\section{Related Work}
Many techniques have been proposed to prevent catastrophic forgetting. In general, these methods regularize the neural net to not deviate too far from previously learned configuration in some sense \cite{kirkpatrick2017overcoming,zenke2017continual,li2017learning,lee2017overcoming,he2018overcoming}. The differences typically lie in how the deviation is measured and how the constraint is enforced.

\citet{kirkpatrick2017overcoming} proposed EWC, which stores the sensitivity of previous task loss to different parameters, and penalizes model parameter changes from one task to the next according to the different sensitivities. 
Further improvements and more sophisticated methods to define parameter change penalties are then proposed by \citet{zenke2017continual, lee2017overcoming,he2018overcoming}. 
GEM~\cite{lopez2017gradient} defines constraints using parameter gradients, so its regularization is also on the model parameters. Practically, however, they either do not significantly outperform EWC on the simplest Permuted MNIST problem \cite{zenke2017continual,lee2017overcoming}, or require prohibitively more storage and computation than EWC \cite{he2018overcoming}. A common deficiency of these approaches is the fact that the constraint on the model is placed on neural net weights, leading to two problems: first, they need additional storage that is at least proportional to the model parameter size, which can be prohibitive because many neural networks involve millions of parameters; second, one ultimately cares about whether the learned functional mapping changes in undesired ways, rather than the weights.

On the contrary, our proposed FSR directly place the regularization on the function mapping instead of parameters. Moreover, it does so with only a small episodic memory of previous data and their corresponding logits, significantly smaller than storing the whole model. 
The recording and re-use of previous samples is similar to experience replay in the Reinforcement Learning (RL) literature \cite{lin1992self,mnih2015human,schaul2015universal,andrychowicz2017hindsight}. However, the goal of experience replay in RL is mainly to break temporal correlation in the input data sequence, whereas our goal here is to avoid catastrophic forgetting. Interestingly, there has been evidence from neuroscience that suggests the brain uses hippocampal replay to consolidate memory \cite{derdikman2010dual,olafsdottir2018role}. 

Learning without Forgetting (LwF)~\cite{li2017learning} is similar to our method in that, the regularization is in function space rather than the parameter space. However, LwF matches the predicted labels of previous models on the \emph{current} data, while our method matches the logits of previous models on the \emph{memory} data. 
Moreover, LwF has two issues: first, when the input distribution changes significantly across tasks, matching current data's outputs may not lead to good performance on the previous data~\cite{rannen2017encoder}; second, it also needs to store the whole model from previous tasks to compute outputs on the current data, leading to prohibitive space requirement. In our experiments, we demonstrate how LwF is unsatisfactory in both memory cost and performance in preventing catastrophic forgetting.

The logit matching used in our FSR approach dates back to early work on model compression~\cite{bucila2006model}. This approach or the similar idea of distillation were used on a large amount of data to compress one or an ensemble of large models into a much more compact model~\cite{ba2014deep,hinton2015distilling}. 
\citet{hinton2015distilling} also argued that logit matching of \citet{bucila2006model} is a special case of distillation. 
In this work, we adopt these ideas, but use them in a very different way: instead of distillation/logit matching on a large amount of input-output data, we apply them on a small subset to obtain the desired behavior on a larger dataset. 

The most closely related method to our proposed FSR is iCaRL~\cite{rebuffi2017icarl}, which also uses distillation to prevent catastrophic forgetting. 
We will compare our method to iCaRL in details in \cref{subsec:compiCaRL} and in the experiments.

\section{Few-Shot Self Reminder}
\label{sec:FSR}
\newcommand{\EE}{\mathbb{E}}
\newcommand*{\argmin}{\mathop{\mathrm{argmin}}}

We introduce our method FSR on the continual learning setting where 
the space of visual concepts is fixed and known, but the input distribution drifts over time. Such a setting has practical implication for almost all real-world production environment for machine vision as newly collected data potentially come from different environmental conditions. For example, a production vision system for traffic sign recognition could face this scenario. The classification outputs are known and fixed a priori, but input data could be gathered incrementally as different road conditions are tested.  Later in \cref{sec_class_increment}, we will discuss how FSR applies to incremental settings where new output visual concepts need to be classified. 

Given a sequence of datasets $\cD_1,\cD_2,\cdots$, one at a time, 
the goal is to attain a model $f_T:\cX\mapsto\cY$ that performs well on the first $T$ datasets after sequentially trained on the $T$ tasks, where  $\cX$ is the input space and $\cY$ is the $K$-dimensional probability simplex. 
The value of $T$ is not known in advance so we would like to have a good model $f_T$ for any $T$ during the sequential training. 
The constraint in the continual learning setting is that at task $T$, we would not be able to access the data from the previous tasks, but can only carry over a limited amount of information about the previous tasks.
This constraint, while more realistic in practice, also excludes a na\"{i}ve solution: simply join all the datasets as one big dataset and train $f_T$ on it in a multi-task learning fashion. 
This learning problem is challenging in that, if we simply re-train the same model over and over using the current available dataset $\cD_T$, it will forget how to properly predict for datasets $\cD_t, t<T$. 
This is known as the catastrophic forgetting problem.

Denote the loss for all the previous datasets by $
L^{(T)}(f) = \sum_{t=1}^T \EE_{\cD_t}\left[ L\left(f(X), Y \right)\right]$,
where $(X,Y)$ is the random data pair in dataset $D_t$ for $1\le t\le T$. For simplicity, we denote $ \EE_{\cD_t}\left[ L\left(f(X), Y \right)\right]$ by $L_{\cD_t}(f)$. 
Let $f_T \defeq \argmin_{f} L^{(T)}(f)$.
Noting that for any sequence of $f_1, \cdots, f_{T-1}$, we can rewrite
\[
\min_f L^{(T)}(f) 
=\sum_{t=1}^{T-1} L_{\cD_t}(f_t)
+ \min_f L_{\cD_T}(f) + \sum_{t=1}^{T-1} 
\Delta_{\cD_t}(f,f_t),
\]
where $\Delta_{\cD_t}(f, f_{t}) = L_{\cD_t}(f) - L_{\cD_t}(f_{t})$ measures the difference in the performances of $f$ and $f_{t}$ on $\cD_t$. 
Given $f_1, \cdots, f_{T-1}$ that are learned from the previous tasks, learning $f_T$ requires minimizing $L_{\cD_T}(f) + \sum_{t=1}^{T-1} \Delta_{\cD_t}(f,f_t)$.

It remains to decide what information from $\cD_t$ is important for $f$ to achieve a small $\Delta_{\cD_t}(f,f_t)$, given a limited amount of memory. 
One immediate observation is that the distance of $\theta$ (parameters of $f$) to $\theta_t$ (parameters of $f_t$) is less concerned, as long as the predicting behaviors of $f$ remain similar to that of $f_t$. 
A more direct approach would be to pass a small number of samples $\widetilde{\cD}_t = \{(x_j^{(t)}, y_j^{(t)})\, \vert \, j=1,\cdots,m\}$ from $\cD_t$ (or the predicted labels), thus $\sum_t \Delta_{\cD_t}(f,f_{t})$ can be replaced by $\sum_t \Delta_{\widetilde{\cD}_t}(f,f_{t})$. 
However, such approach depends heavily on that a single label can represent the structured output of $f_t$, which is unrealistic in general.
In order to pass the information of the structured output to fully reproduce the predicting behavior of a model, we propose our Few-shot Self Reminder (FSR) based on the following ``self-distillation'' method on the logits:
\begin{equation}
\label{eq:matchinglogits}
\min_{\theta}\ 
\frac{1}{n_T}\sum_{i=1}^{n_T} L(f(x^{(T)}_i),y^{(T)}_i) 
+\frac{\lambda}{m}\sum_{t=1}^{T-1}\sum_{j=1}^m 
\|z_j - z^{(t)}_j\|_2^2,
\end{equation}
where $n_T$ is the number of samples for task $T$, $\lambda$ is a regularization parameter, and $z_j,z^{(t)}_j$ are the logits of the memory data $x_j$ produced by $f$ and $f_t$ respectively.

Intuitively, the selected points $\widetilde{\cD}_t$ should be representative and provide as much constraint as possible to change in $f$. Surprisingly, it turns out that class-stratified random sampling already works exceptionally well in our experiments. 
We also test out an efficient parameter-gradient based estimation method for finding representative points. The intuition here is that representative points are both easier to learn (comparing to ``corner cases'') and occur more frequently in the training set. Hence as the initial transient phase of learning epochs, representative points should contribute less model parameter gradient on average. 
Empirically, this method sometimes outperforms stratified random sampling, but not always with a significant margin. See Supplementary for more details. Hence, the simplest class-stratified sampling is still the most preferred selection method in the end.

There has been some attempt at designing submodular scoring functions for subset selection in continual learning~\cite{Brahma_2018_CVPR_Workshops}. However, as the authors showed in the preliminary work, it does not significantly outperform class balanced random sampling. Other sophisticated methods such as influence function~\cite{koh2017understanding} and kernel herding~\cite{chen2010super} could potentially be used for point selection. But we found that they are noticeably worse than random sampling in this problem. Results can be found in the Supplementary. 

In the following, we will discuss several alternative approaches and how they differ from our FSR with logit matching.
\cref{sec_class_increment} will then provide an extension to handle new visual concepts.

\subsection{Matching Logits vs. Matching Probabilities}
FSR could alternatively match the \emph{outputs} of $f$ and $f_t$ on $\widetilde{\cD}_t$.  
Such idea leads to the following ``self-distillation'' method on the soft labels:
\[
\frac{1}{n_T}\sum_{i=1}^{n_T} L(f(x^{(T)}_i), y^{(T)}_i) 
+ \frac{\lambda}{m}\sum_{t=1}^{T-1}\sum_{j=1}^m \KL(f_t(x^{(t)}_j)\|f(x^{(t)}_j)),
\]
where $\KL$ is the KL divergence. 
Here the regularization term resembles model distillation~\cite{hinton2015distilling}, which is originally proposed to solve the model compression problem.  
Given two distributions $y$ and $y'$, that are the outputs of the softmax function on the logits $z$ and $z'$. 
Note that
\begin{equation}
\begin{split}
\KL(y \, \| \, y') 
&= \sum_k y_k\left( \log y_k - \log y_k'\right) \\
&= \sum_k y_k\left( z_k/\tau - z'_k/\tau \right) + \left(Z - Z' \right), \end{split}
\label{eq:distillation}
\end{equation}
where $\tau$ is the temperature hyperparameter, and $ Z  = \log \sum_k \exp\left( z_k/\tau \right)$ is the normalizer for the softmax function. 

One immediate observation is that the softmax function is invariant in constant shift in its logits, i.e. $\text{softmax}(z) = \text{softmax}(z-c)$ for any constant $c$, thus matching logits is a stronger requirement compared to matching probability output.
Furthermore, assuming that $Z = Z'$, \cref{eq:distillation} can be interpreted as a \emph{weighted} sum of the logits mismatches. 
Discrepancies in small probabilities are ignored in distillation, even if their relative ratio could be very large, which conveys very important information about the underlying function (a.k.a. ``dark knowledge''). 
\citet{hinton2015distilling} further propose to use a large temperature for distillation so that the regularizer would not focus only on the predicted label\footnote{Note that for a confident model which assigns $y_k$ close to 1 for some $k$, the summation over all $k$ boils down to a single term on the predicted label.}. 
Compared to distillation in \cref{eq:distillation}, matching logits in \cref{eq:matchinglogits} places \emph{equal} weights on all the logits, which automatically solves the above ``winner-takes-all'' problem.

\subsection{Compared to iCaRL}
\label{subsec:compiCaRL}
iCaRL~\cite{rebuffi2017icarl} focuses on the class-incremental learning and uses a memory of exemplars as class prototypes.
Because of this setting, iCaRL uses independent logistics in the outputs for representation learning. There are issues associated with using different objectives for representation learning and classifier learning, as well as combining different objectives in an ad hoc fashion. 
This problem of ad hoc objectives is not the focus of this work. 

Although iCaRL is proposed in a very different learning setting, its self-distillation part is closely related to our method. In particular, instead of a $L_2$ distance over the logits, iCaRL uses the following regularizer
\[
\frac{\lambda}{m}\sum_{j=1}^m \sum_{k=1}^{K} \text{CE}(g(z_{j,k}^{(T-1)}), g(z_{j,k})),
\]
where $g(z) = \frac{1}{1 + \exp(-z)}$ is the logistic function, $\text{CE}(p,q)$ is the cross entropy function defined as $\text{CE}(p,q) = -p\log q - (1-p) \log (1-q)$, and $z_{j,k},z_{j,k}^{(T-1)}$ are the $k$th logits of sample $x_j$ under $f,f_{T-1}$ respectively.
Since $g$ is invertible, iCaRL also tries to match the logits in the ideal situation. 
However, compared to our method, iCaRL suffers the problem of losing ``dark knowledge''. 
Such phenomenon may be more obvious when checking the gradient (learning signal) provided by this regularization term. 
By simple algebraic calculation, the gradient of this regularizer w.r.t. the logit is 
\[
\frac{\lambda}{m}\sum_{j=1}^m \sum_{k=1}^{K} 
g(z_{j,k}) - g(z_{j,k}^{(T-1)}). 
\]  
First, iCaRL ignores the mismatch when $z_{j,k}$ and $z_{j,k}^{(T-1)}$ are large/small yet significantly different. Second, given $z_{j,k}$ and $z_{i,l}$ are in a proper magnitude, iCaRL assigns almost the same learning signal for both logits at $(j,k)$ and $(i,l)$ when $z_{j,k}^{(T-1)}$ and $z_{i,l}^{(T-1)}$ are both large/small yet significantly different. In a nutshell, iCaRL loses some information when the logits reside in the saturating areas of the logistic function, and our FSR behaves qualitatively different from the distillation used in iCaRL.

\subsection{Compared to EWC}

Suppose that $f$ is parametrized by $\theta$, and $\theta_t$ is the parameter of $f_t$ for $t = 1,\cdots,T$. We may also use $f_{\theta_t}$ as $f_t$ to emphasize its dependence on $\theta_t$ in the following. 
Different from our approach, EWC~\cite{kirkpatrick2017overcoming} handles this problem in a recurrent way. 
For any $f_{T-1}$, note that 
\begin{align*}
L^{(T)}(f) = L_{\cD_T}(f) + \sum_{t=1}^{T-1} [\Delta_{\cD_t}(f, f_{T-1}) +  L_{\cD_t}(f_{T-1})].
\end{align*}
EWC incrementally learns $f_T$ from $f_{T-1}$:
\[
\min_f \quad L_{\cD_T}(f) + 
\sum_{t=1}^{T-1} \Delta_{\cD_t}(f,f_{T-1}).
\]
Assuming that $f_{T-1}$ has performed well on all the previous data $\cD_1, \cdots, \cD_{T-1}$, 
$\sum_t\Delta_{\cD_t}(f, f_{T-1})$ is approximated by a second order approximation when the loss function $L$ is the KL divergence $\KL$:
\[
\min_{\theta}\quad
\frac{1}{n_T}\sum_{i=1}^{n_T} 
L(f_{\theta}(x^{(T)}_i), y^{(T)}_i)
+ \frac{\lambda}{2}\|\theta-\theta_{T-1}\|_{F_{T-1}}^2,
\]
where $\|\cdot\|_{F_{T-1}}^2$ is the norm defined by the Fisher information matrix (derived in Supplementary). 
This approximation has good accuracy when $\theta$ is closed to $\theta_{T-1}$. 
In practice, $F_{T-1}$ is further approximated by $\sum_{t=1}^{T-1} \mathcal{H}_\theta\left(L_{\cD_t}(f_{t})\right)$, where $\mathcal{H}_\theta(\cdot)$ is the Hessian. 

Although the idea of EWC seems natural and technically sound, it suffers several problems in practice. 
The first problem is that the memory cost of EWC is $O(p)$ as it needs to store previous model's parameters $\theta$ and the corresponding Fisher, where $p$ is the number of parameters of the model. 
This memory cost could be demanding since nowadays many deep neural networks involve millions of parameters.
More importantly, by regularizing the distance between $\theta$ and $\theta_{T-1}$, EWC makes an implicit assumption that the tasks are highly related to each other. Therefore the learning would only update $\theta$ in a ``small'' range for all the tasks. 
When this assumption is violated, the approximation on the Fisher matrix may not be accurate. 
In particular, the Fisher matrix is \emph{fixed} rather than being evaluated on the current $\theta$ during the training on the $T$th task. 
Furthermore, $\mathcal{H}_\theta\left( L_{\cD_t} (f_{\theta_{t}})\right)$ may not be a good approximation to $ \mathcal{H}_\theta\left(L_{\cD_t} (f_{\theta_{T-1}})\right)$ when $t\ll T-1$. 
Note that other gradient based methods like GEM~\cite{lopez2017gradient} also share such problems.

\subsection{Tasks with New Visual Concepts}
\label{sec_class_increment}
So far we have focused our discussion on the setting where input distribution changes between tasks, but with fixed output space. 
Now we consider situations where new output visual concepts need to be classified. 

Suppose that each task $t$ consists of data $\cD_t$ from input/output space pair $(\cX,\cY_t)$, where the dimensions of $\cY_t$ do not correspond to the same classes across different tasks. In fact, the number of dimensions of $\cY_t$ could differ. This setting corresponds to situation where new sets of visual concepts need to learned. For instance, the sequence of tasks could consist of fine-grained classifications of different dogs, different birds, different cars, etc. We refer to this setup as {\em class-incremental}.

We will use models with {\em multiple heads} for this setup, where a different set of output layers is added for each new task, and all layers below are shared. FSR applies in the most straightforward way in this case where logit matching loss is enforced between a corresponding old recorded logit and new evaluation at the old task's head.

\section{Experiments}
\label{sec_exp}
In this section, we empirically demonstrate that our proposed FSR approach forgets much slower than popular alternatives and that it can handle highly dissimilar tasks. 
We use two variants of our proposed FSR approach: logit matching (labeled as Logit) and distillation (labeled as Distill, as a reference method). 
The baselines are vanilla SGD, LwF~\cite{li2017learning}, iCaRL~\cite{rebuffi2017icarl}, EWC~\cite{kirkpatrick2017overcoming} and HAT~\cite{serra2018overcoming}. 
SGD is a na\"{i}ve baseline used to showcase performance of EWC by \citet{kirkpatrick2017overcoming} so we include it as comparison.
LwF and iCaRL share some similarity to our method, but have important distinctions as outlined in previous sections. 
We also compare to HAT, which is recently shown to be superior over popular alternatives in the class-incremental setting. 
The regularization parameter of each method is individually tuned with a large range of candidates, based on a hold-out validation partition. 
Unless specified otherwise, SGD uses a step size of $0.001$, and all other methods are trained using the Adam optimizer~\cite{kingma2014adam} with step size of 0.0001. 
Supplementary provides further experiment details.

We will first introduce the the benchmark problems. 
Because our FSR can trade-off between the memory usage and knowledge retention. 
Our analysis will then first demonstrate how FSR can forget slower than existing methods when using comparable memory (\cref{sec:little_forgetting}), followed by several surprising results when applying FSR with very small memories (\cref{sec:little_memory}). 
Finally, \cref{sec:inc_cifar100} illustrates how FSR performs in the class-incremental setting.

\textbf{Permuted MNIST} 
The first setting is the permuted MNIST problem~\cite{goodfellow2013maxout,lecun1998gradient}, a popular benchmark for continual learning~\cite{kirkpatrick2017overcoming,zenke2017continual,lopez2017gradient}. 
For each task, a fixed random permutation is applied to all input images. 
Note that pixel permutation is a linear transformation, thus the resulting tasks are relatively similar to each other. 
We select 500 class-balanced MNIST images per task as memory for FSR, which in total is comparable to the memory cost of EWC (see Supplementary). 

\textbf{Nonlinearly Transformed MNIST} 
To compare how different methods handle tasks that are less similar, we further design a nonlinearly transformed MNIST benchmark. 
In this problem, a fixed random nonlinear (but invertible) transformation is applied to all the images. 
Specifically, we use a four-layer fully connected MLP as a transformer. 
Its weights are randomly initialized with orthogonal initialization.
All layers have the same number of units (784) and use Leaky ReLU ($\alpha=0.2$) as activation. 
The output image is finally re-normalized to the [0,1] range.
Each task corresponds to a set of different randomly initialized weights. 
Since the transformation is invertible, no information is lost, which ensures that each task is still equally solvable by a permutation invariant model like MLP. 

\textbf{Color Space Transformed CIFAR10} 
To test the methods on natural images, we conduct experiments with color space transformations of the CIFAR10 dataset~\cite{krizhevsky2009learning}. Here, the transformation is applied in the channel dimension on the pixel level and we use different color space encodings as different tasks. 
The original CIFAR10 is based on RGB encoding. 
The five color spaces used in the experiments are RGB, YIQ, YUV, HSV, HED (order as listed)~\footnote{Implemented using the scikit-image library: scikit-image.org}. 
The YIQ and YUV spaces are linear transformations of the RGB space, while HSV and HED are nonlinear transformations. 
This ordering ensures that the tasks are getting progressively harder and forgetting is more aggravated if not handled well. 
A VGG-like model with enough hidden units (see Supplementary) is used for this learning task to accommodate different color space inputs. 
3000 class-balanced images are chosen from each task as memory for FSR, which in total is comparable to the memory usage of EWC. 

\textbf{Incremental CIFAR100} 
In this problem, the CIFAR100 dataset~\cite{krizhevsky2009learning} is partitioned into $10$ disjoint sets of $10$ classes, and we train an AlexNet with multiple distinct output layers on the $10$ tasks sequentially. Each run corresponds to a new partitioning. We follow the class-incremental protocol in \citet{serra2018overcoming} and report the forgetting ratio during the training. 
To use a comparable memory as EWC, FSR stores 400 class-balanced images per task. 
Here, all methods use SGD with a decaying step size initialized to be 0.05.

\begin{figure}[h]
\centering
  \begin{subfigure}[b]{\FigWidth}
  \includegraphics[width=.95\textwidth]{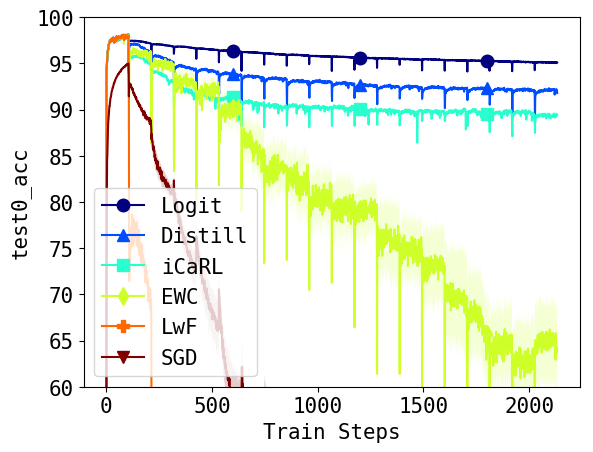}
    \caption{First Task (Mean \& standard error over 5 runs)}
    \label{subfig:permute_test0_acc}
  \end{subfigure}%
  \hfill
  \begin{subfigure}[b]{\FigWidth}
    \includegraphics[width=.95\textwidth]{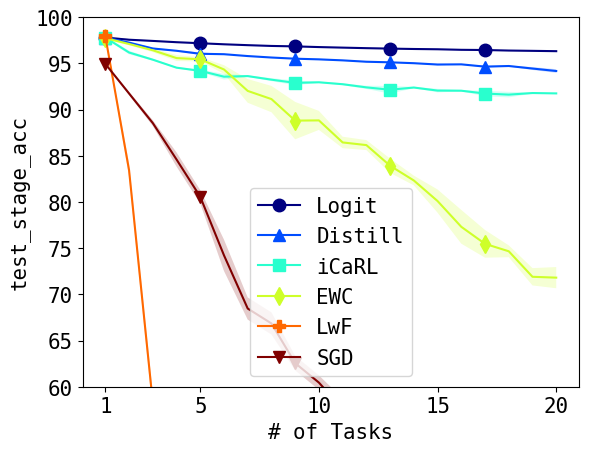}
    \caption{Task Average (Mean \& standard error over 5 runs)}
    \label{subfig:permute_test_stage_acc}
  \end{subfigure}
  \caption{Permuted MNIST Test Accuracy}
  \label{fig:permute_test_acc}
\end{figure}

\subsection{Little Forgetting}
\label{sec:little_forgetting}

\textbf{Permuted MNIST} We use a five-layer MLP for this setting with 1024 hidden units except the last layer, which has 10 units. 
We use a richer model than those in the prior works as we will use the same model for learning nonlinearly transformed MNIST later, which is a significantly more challenging problem. 

\cref{fig:permute_test_acc} shows the results. 
\cref{subfig:permute_test0_acc} shows the test accuracy of the first task along with the training of 20 sequential tasks, while \cref{subfig:permute_test_stage_acc} shows the average test accuracy of tasks thus far. 
We can see that all methods except LwF outperform SGD with a large margin. 
LwF performs poorly in this problem due to two possible factors: (1)~noticeable distribution changes in the input space, as also pointed out by other researchers~\cite{rannen2017encoder} and (2)~the fact that the two losses based on ground truth label and distilled label from previous task are in fact conflict with each other. 
It is difficult for a single model to predict both labels given the same image. 
Another observation from \cref{fig:permute_test_acc} is that, matching logits, distillation, and iCaRL have a significant improvement over EWC when using comparable memory size. 
EWC performs reasonably well for the first 5 tasks, after which it degrades severely. 
We also tried to compare to Generative Replay~\cite{shin2017continual}. 
However, we are not able to reproduce their results without important experiment details, but as a reference, it achieved about 95\% of average accuracy after 5 permuted MNIST tasks~\cite[Fig.2b]{shin2017continual}~\cite[Table 7]{serra2018overcoming}, which is worst than FSR as shown in \cref{fig:permute_test_acc}.

\begin{figure}[h]
\centering
  \begin{subfigure}[b]{\FigWidth}
    \includegraphics[width=.95\textwidth]{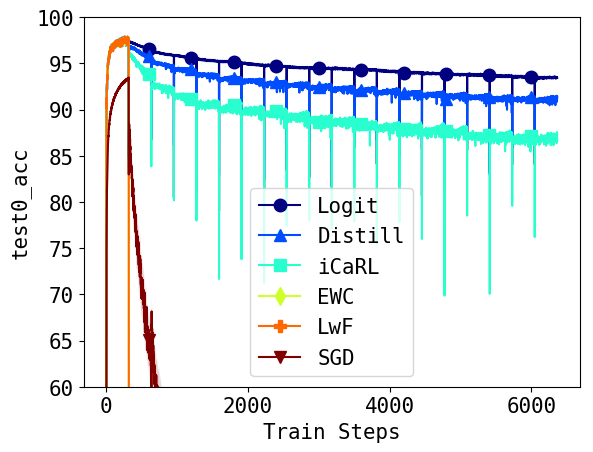}
    \caption{First Task (Mean \& standard error over 5 runs)}
    \label{subfig:nonlinear_test0_acc}
  \end{subfigure}%
  \hfill
  \begin{subfigure}[b]{\FigWidth}
    \includegraphics[width=.95\textwidth]{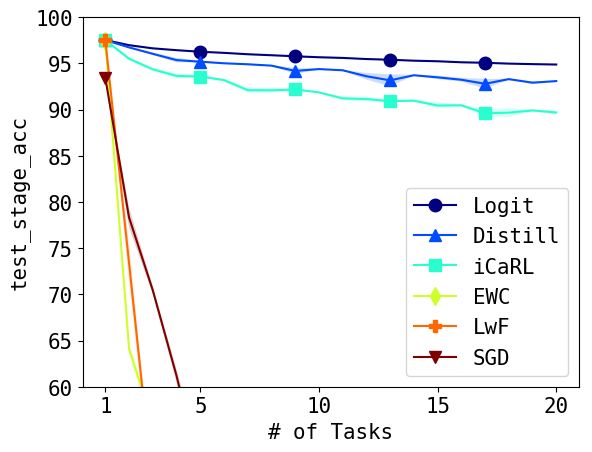}
    \caption{Task Average (Mean \& standard error over 5 runs)}
    \label{subfig:nonlinear_test_stage_acc}
  \end{subfigure}
  \caption{Non-Linear MNIST Test Accuracy}
  \label{fig:nonlinear_test_acc}
\end{figure}

\textbf{Nonlinearly Transformed MNIST}
We then test on the a more challenging problem of nonlinearly transformed MNIST, where tasks are less similar.
The results are shown in \cref{fig:nonlinear_test_acc}. 
As we anticipated, when data distributions are much different from task to task, approaches that match model parameters like EWC can fail miserably. 
Essentially, EWC only utilizes local information as the diagonal Fisher matrix. 
When the optimal solutions of two tasks are far apart, the local information of the first task is no longer accurate during the training process of the second task, and there might not be overlap for the two estimated Gaussian ellipsoids. 
On the contrary, methods that solely match the outputs of previous models like FSR can still maintain a remarkably better performance than EWC. 

\begin{figure}[h]
\centering
  \begin{subfigure}[b]{\FigWidth}
    \includegraphics[width=.95\textwidth]{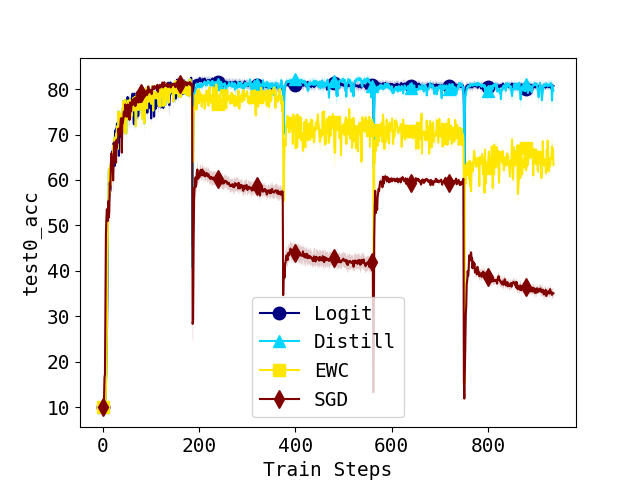}
    \caption{First Task (Mean \& standard error over 5 runs)}
    \label{subfig:cifar10_test0_acc}
  \end{subfigure}%
  \hfill
  \begin{subfigure}[b]{\FigWidth}
    \includegraphics[width=.95\textwidth]{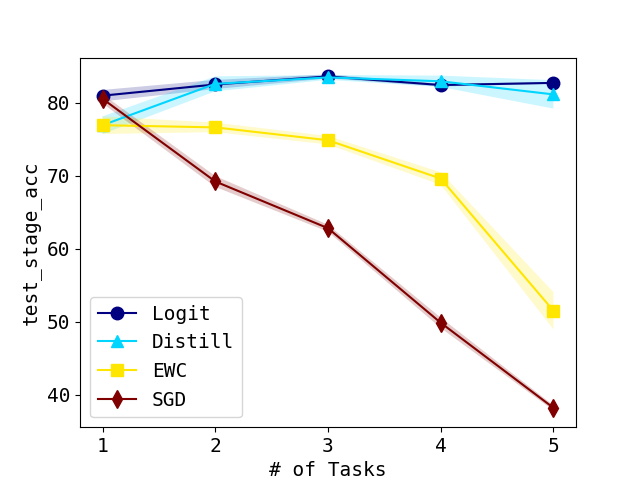}
    \caption{Task Average (Mean \& standard error over 5 runs)}
    \label{subfig:cifar10_test_stage_acc}
  \end{subfigure}
  \caption{CIFAR10 Test Accuracy}
  \label{fig:cifar10_test_acc}
\end{figure}

\textbf{Color Space Transformed CIFAR10}
The results are shown in \cref{fig:cifar10_test_acc}. 
It can be seen that without considering previous tasks, SGD forgets quickly as the model encounter images represented in a new color space. 
EWC can maintain a reasonably good overall accuracy when the transformation is linear, but when the transformation becomes nonlinear, its accuracy drops significantly.
Meanwhile, FSR can preserve or even improve average test accuracy.

\begin{figure}[h]
\centering
  \includegraphics[width=0.85\columnwidth]{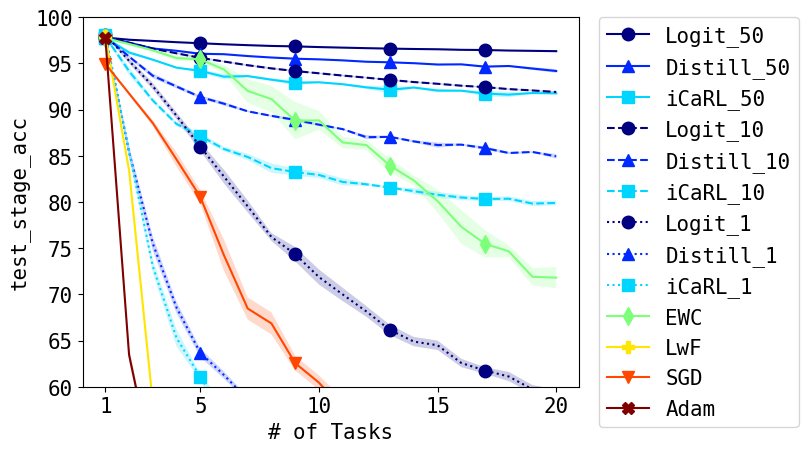}
  \caption{Permuted MNIST average task test accuracy with different memory sizes (mean and standard error over 5 repetitions). Solid lines are using full memory. Dashed and dotted lines are with partial memory. The numbers in the legend indicate the numbers of examples per class per task.}
  \label{fig:mem_test_stage_acc}
\end{figure}

\subsection{Little Memory}
\label{sec:little_memory}

To further examine the effectiveness of FSR, we test our method with small memory. 
FSR can surprisingly do well in the extreme setting of retaining only a few images. 
We focus on the permuted MNIST setting and show the effect of different memory sizes in \cref{fig:mem_test_stage_acc}.
There are a few interesting observations, as we will discuss below.

First, more aggressive optimizer like Adam tends to forget more quickly than vanilla SGD. 
It may be explained by the fact that adaptive optimizers usually find local optimum of the new task quicker than SGD, hence moving away from previous solutions more quickly. 
However, the exact reason for this phenomenon is unknown to the best of our knowledge and outside the scope of this paper.

\begin{figure}[h]
  \centering
  \begin{subfigure}[b]{\TwoColTwoSubfigWidth}
    \includegraphics[width=\textwidth]{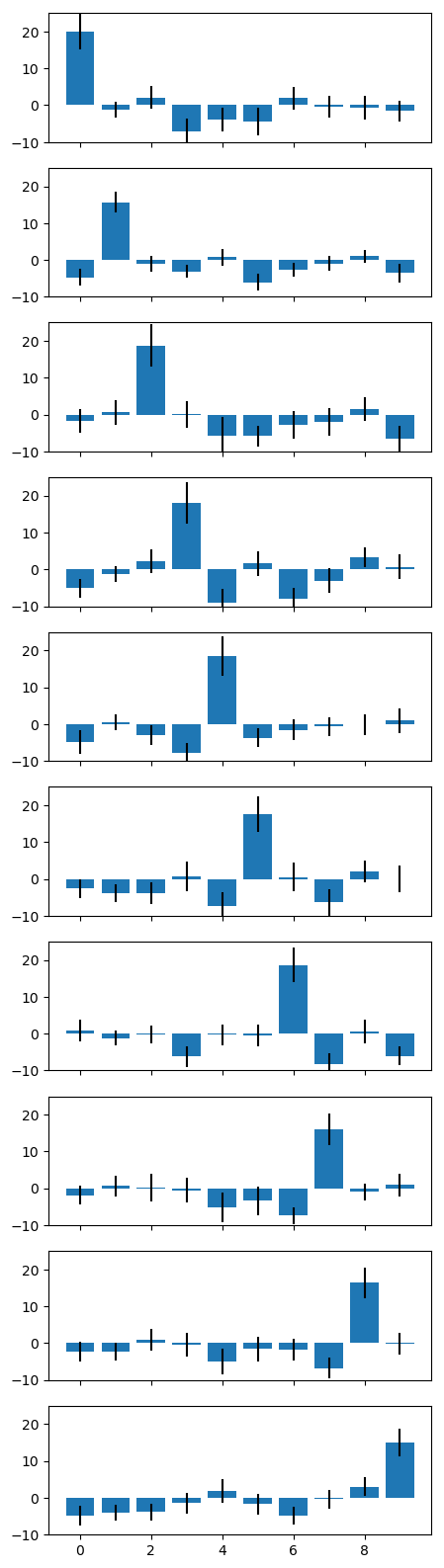}
    \caption{Original}
    \label{subfig:original_logits_bars}
  \end{subfigure}
  \begin{subfigure}[b]{\TwoColTwoSubfigWidth}
    \includegraphics[width=\textwidth]{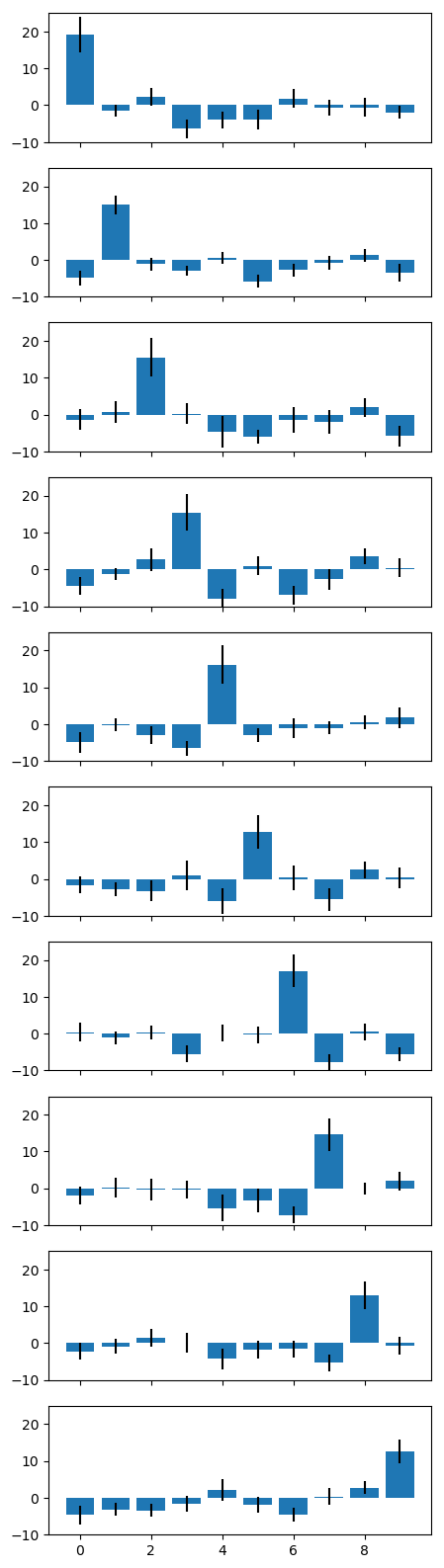}
    \caption{10 per class}
    \label{subfig:10_logits_bars}
  \end{subfigure}
  \caption{Means and Standard Deviations of Logits}
  \label{fig:logits_bars}
\end{figure}

Second, strikingly, even with only 1 image per class (a memory size of 10 images per task), matching logits can improve over SGD by a noticeable margin. 
Recall that we match logits with the Adam optimizer, which means that even with only 1 image per class can remedy the forgetting issue of Adam. 

Third, with 10 images per class (thus 100 images per task), matching logits can outperform EWC for this problem. 
It is surprising that matching logits can perform so well, provided that it only uses $100/500=20\%$ of the memory cost of EWC. 
If we consider only the first 5 tasks, then FSR achieves the performance of EWC with only 5\% of its memory. 
To better understand the effectiveness of FSR, the logits distributions of each MNIST class are provided in \cref{fig:logits_bars}. 
\cref{subfig:original_logits_bars} shows the average logits of images of each class in the hold-out validation partition, after training on the first task.
The first subplot in \cref{subfig:original_logits_bars} shows the average logits of images with label `0', together with their standard deviations as error bars. 
The rest of the subplots are similarly defined. 
Clearly, the model has successfully distinguished between different classes by making the correct labels' logits much higher than those of the incorrect labels. 
\cref{subfig:10_logits_bars} shows the same (of first task validation data) after training on the second task with 10 images per class as memory. 
Even with such small memory, matching logits can generalize very well for unseen data in the first task, which explains why it could be more favorable when the memory budget is tight. 

\begin{figure}[h]
  \centering
  \begin{subfigure}[b]{\TwoColTwoSubfigWidth}
    \includegraphics[width=\textwidth]{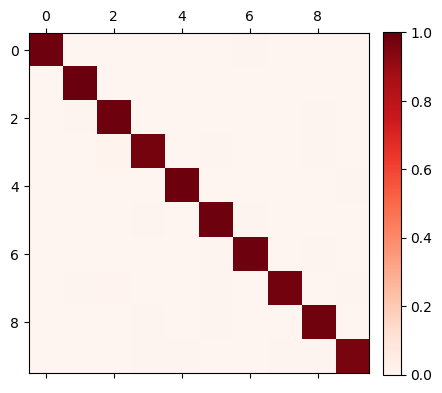}
    \caption{Original}
    \label{subfig:pred_original}
  \end{subfigure}
  \begin{subfigure}[b]{\TwoColTwoSubfigWidth}
    \includegraphics[width=\textwidth]{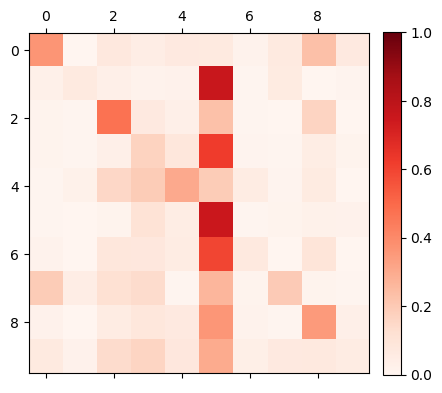}
    \caption{Forgetting}
    \label{subfig:pred_forgetting}
  \end{subfigure}

  \begin{subfigure}[b]{\TwoColTwoSubfigWidth}
    \includegraphics[width=\textwidth]{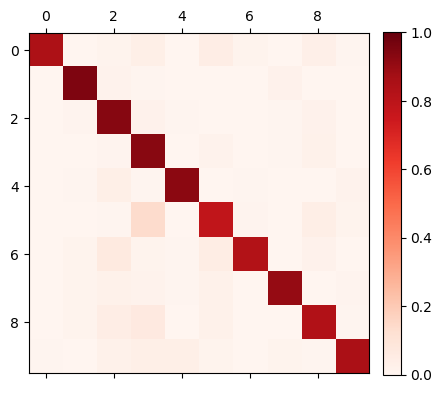}
    \caption{Logits: 1 per class}
    \label{subfig:pred_logits1}
  \end{subfigure}
  \begin{subfigure}[b]{\TwoColTwoSubfigWidth}
    \includegraphics[width=\textwidth]{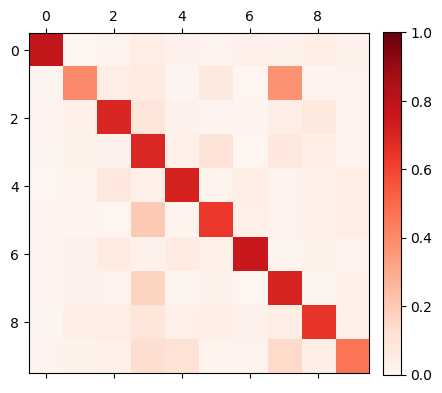}
    \caption{Distill: 1 per class}
    \label{subfig:pred_distill1}
  \end{subfigure}
  \caption{Prediction Heatmaps}
  \label{fig:permute_pred_heatmaps}
\end{figure}

Fourth, as shown in \cref{fig:mem_test_stage_acc}, matching logits consistently performs better than distillation and iCaRL, across all memory sizes.
Their accuracy differences are more significant with smaller memories. 
To see why matching logits is more effective, we have shown the prediction heatmaps in \cref{fig:permute_pred_heatmaps}. 
In each subplot, each row shows the average probabilities of the corresponding class images. 
For instance, the first row is the average predicted probabilities of images of class `0' in the validation partition after training on the first task. 
Using Adam, the model forgets what the predictions of the first task data should be after training on the second task, as shown in \cref{subfig:pred_forgetting}. 
With only 1 single image per class, \cref{subfig:pred_logits1} shows how logit matching manages to generalize well in terms of the prediction probabilities on the validation set. 
On the contrary, distillation is less effective when the memory is small, as in \cref{subfig:pred_distill1}.

\subsection{Incremental Learning}
\label{sec:inc_cifar100}
To showcase FSR in incremental settings, where new task consists of new classes, we compare it with HAT and EWC in the incremental CIFAR100 dataset. 
We report the average forgetting ratio $\rho$, which is a quantity introduced by \citet{serra2018overcoming} to show how much an algorithm forgets comparing to multi-task joint training.
$\rho\approx 0$ indicates no forgetting, while $\rho\approx -1$ means total forgetting. 

\begin{figure}[ht]
\centering
  \includegraphics[width=0.8\columnwidth]{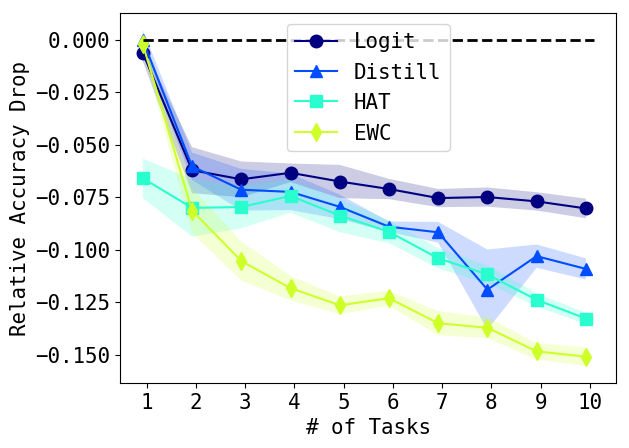}
  \caption{Average Forgetting Ratio for CIFAR100}
  \label{fig:cifar100_rho}
\end{figure}

The results are show in \cref{fig:cifar100_rho}. 
It can be seen that FSR with logit matching outperforms the other alternatives. 
Note that although HAT uses a smaller memory than FSR and EWC, its attention/gating mechanism makes it incapable of sharing important intermediary features common to the tasks. For instance, if one task already contains an insect class and the model has learned features for insect legs, then subsequent stages that have other insect classes would have to relearn features for legs using other parts of the model. This defies the goal of distributed representation reusing features to handle the combinatorially exploding visual world.
Therefore, when the tasks have a lot of shared knowledge like we have in the case of incremental CIFAR100, HAT performs significantly worse than Logit. 
This difference in performance is accentuated when model capacity is limited (see Supplementary). 
We further notice that if the true labels of memory data are also stored and the corresponding KL divergence is added to the objective, FSR can perform even better (see Supplementary). 
Despite our best effort of tuning hyperparameters, HAT \mbox{cannot} achieve $\rho\approx 0$ for the first task. 
On the other hand, for setups where there are less sharing, HAT performs better than FSR, like in the setup where $8$ different classification datesets are taken to be the sequence of tasks, such as MNIST, CIFAR~\cite{krizhevsky2009learning}, SVHN~\cite{netzer2011reading}, FaceScrub~\cite{ng2014data}, etc.

\section{Conclusions}
To overcome the catastrophic forgetting problem in continual learning of deep neural networks, we proposed Few-shot Self Reminder~(FSR) that requires substantially smaller memory yet forgets slower than popular alternatives. As a side contribution, we also introduced a new and harder benchmark, nonlinearly transformed MNIST, with more substantial between-task dissimilarities.

Finally for point selection, although both our and other attempts in the literature find class-stratified sampling to very effective, it would be useful to further investigate the parallels to GP sparsification techniques~\cite{herbrich2003fast,cao2013efficient}, where the goal is also to select points which provide significant information about the structure of decision boundaries.

{\small
\setlength{\bibsep}{0pt}
\bibliographystyle{abbrvnat}
\bibliography{ref.bib}
}

\clearpage
\appendix
\newcommand{\floatThirtyTwo}{\texttt{float32}}
\newcommand{\uintEight}{\texttt{uint8}}

\section{Fisher $L_2$ distance as a second order approximation}
\label{app:fisher}

By definition, the total Fisher information matrix up to time $T-1$ is the sum of individual Hessians:
\[
F_{T-1} = \mathcal{H}_\theta
\left( \sum_{t=1}^{T-1}L_{\cD_t}(f_{T-1})\middle)
\right|_{\theta=\theta_{T-1}}.
\]
Note that $\nabla_{\theta} \sum_{t=1}^{T-1}L_{\cD_t}(f_{T-1}) =0$ by definition of $f_{T-1}$. 
Thus, by second-order Taylor expansion,
\begin{align*}
&\sum_{t=1}^{T-1}\Delta_{\cD_t}(f, f_{T-1})
=\sum_{t=1}^{T-1}L_{\cD_t}(f) - \sum_{t=1}^{T-1}L_{\cD_t}(f_{T-1}) \\
&\approx (\theta - \theta_{T-1})^{\top} \mathcal{H}_\theta
\left( \sum_{t=1}^{T-1}L_{\cD_t}(f_{T-1})\middle)
\right|_{\theta=\theta_{T-1}} 
(\theta - \theta_{T-1}) \\
&= \|\theta-\theta_{T-1}\|_{F_{T-1}}^2.
\end{align*}

\section{Experiment Details}
\label{app:exp_details}
\begin{itemize}
\item \# of epochs: 20 (permuted MNIST), 60 (non-linear MNIST), 40 (CIFAR10).
\item Batch size 128
\item Weight decay 0.0001
\item 5 runs/repititions
\item Distillation temperature $\tau=2$ as chosen in the original papers
\item Regularization parameters $\lambda$
\begin{itemize}
  \item Permuted MNIST: 
The best regularization hyperparameters for Logit, Distill, iCaRL, EWC, LwF are 5, 10, 20, 400, 1 respectively. 
  \item Non-linear MNIST: The best regularization hyperparameters of Logits, Distill, EWC, LwF found on holdout validation sets are 1, 10, 20, 10, 1 respectively.
\end{itemize}
\item Additional for CIFAR10: batch normalization to speed up training
\end{itemize}

\section{Memory Computation}
\label{app:memory}

The following computation of number of model parameters ignores the biases for simplicity.

\textbf{MNIST}. 
The model is five-layer fully connected MLP: 
\begin{align*}
28\times 28
&\xrightarrow{784\times 1024}&
1024
&\xrightarrow{1024\times 1024}&
1024
&\xrightarrow{1024\times 1024}& \\
1024
&\xrightarrow{1024\times 1024}&
1024
&\xrightarrow{1024\times 10}&
10
&&
\end{align*}
The total number of parameters is 3,958,784. 
However, EWC requires another set to store the diagonal of the Fisher, so in total there are 7,917,568 {\floatThirtyTwo} numbers.
Each MNIST image is of size $28\times 28+10=794$ where the 10 is for its output logits/probs.
Therefore, for 20 tasks, each can have $7917568/794/20\approx 500$ images.
Note that the original MNIST format is based on {\uintEight} instead of {\floatThirtyTwo} for the images, which means we can in fact store much more images if the memory is also based on {\uintEight} for the images. 

\textbf{CIFAR10}. 
The model is VGG-like ``ccpccpccpff'', where `c' means convolution, `p' means $2\times 2$ max-pooling and `f' means fully connected:
\[
\begin{split}
&32\times 32\times 3 
\xrightarrow{\text{c:}5\times 5}
32\times 32\times 128
\xrightarrow{\text{c:}5\times 5}
32\times 32\times 128
\xrightarrow{\text{p}}\\
&16\times 16\times 128
\xrightarrow{\text{c:}5\times 5}
16\times 16\times 256
\xrightarrow{\text{c:}5\times 5}
16\times 16\times 256
\xrightarrow{\text{p}}\\
&8\times 8\times 256
\xrightarrow{\text{c:}3\times 3}
8\times 8\times 512
\xrightarrow{\text{c:}3\times 3}
8\times 8\times 512
\xrightarrow{\text{p}}\\
&4\times 4\times 512
\xrightarrow{\text{c:}3\times 3}
4\times 4\times 1024
\xrightarrow{\text{c:}3\times 3}
4\times 4\times 1024
\xrightarrow{\text{p}}\\
&2\times 2\times 1024
\xrightarrow{\text{f:}4096\times 1024}
1024
\xrightarrow{\text{f:}1024\times 10}
10
\end{split}
\]
The parameters involved are
\begin{align*}
\text{ccp:}&&
5\times 5\times 3\times 128
\quad &&
5\times 5\times 128\times 128\\
\text{ccp:}&&
5\times 5\times 128\times 256
\quad &&
5\times 5\times 256\times 256\\
\text{ccp:}&&
3\times 3\times 256\times 512
\quad &&
3\times 3\times 512\times 512\\
\text{ccp:}&&
3\times 3\times 512\times 1024
\quad &&
3\times 3\times 1024\times 1024\\
\text{ff:}&&
4096\times 1024
\quad &&
1024\times 10\\
\end{align*}
In total, there are 24,776,064 {\floatThirtyTwo} parameters. 
However, taking into account that we need another set to store the diagonal of the Fisher, the total memory for EWC is 49,552,128.
Each CIFAR10 image is of size $32\times 32\times 3+10=3082$ where the 10 is for its output logits/probs.
Therefore, for 5 tasks, each can have $49552128/3082/5\approx 3216$ images.
To make things easier, we store 3000 images per task.

\textbf{CIFAR100}.
For this problem, we use the multi-head AlexNet model provided by \citet{serra2018overcoming}, which consists of 6.8M parameters including bias parameters.
Thus EWC uses a memory of size 13.6M, which results in $13.6M/3082/10\approx 422$ images. 
In the experiment, we store 400 images per task.

\section{How to Select Points?}
\label{app:select_points}
In this section, we provide experiment results on different methods for selecting data points into memory.
Although we have tried several other methods as discussed below, eventually class-stratified random sampling still outperforms the alternatives.

Class-stratified random sampling (denoted as Rand in the following) guarantees that each class will have some representatives so that different patterns of logits can be equally seen by later training. 
We also test several other well-known methods such as influence function~(Infl), kernel herding~(Herd) and our proposed sampling based on parameter gradients~(Grad).
In the experiments, all these methods are class-stratified.
To use influence function, we try to select the training data points that have the most positive (increasing validation loss) and most negative (decreasing validation loss) influence.
Points with negative influence are better so we report it in the following results.
As for herding, we tried linear kernel and Gaussian kernel with tuned kernel width.
Linear kernel in fact performs better so we report it in the following.
In our implementation, we already remove the possibility of choosing duplicates during the herding procedure and we use the typical Frank-Wolfe step size $O(t^{-1})$ to ensure that all selected points have equal weights.
The gradient-based method is as described in \cref{sec:FSR}.

\begin{figure}[h]
\centering
  \begin{subfigure}[b]{.65\columnwidth}
  \includegraphics[width=\textwidth]{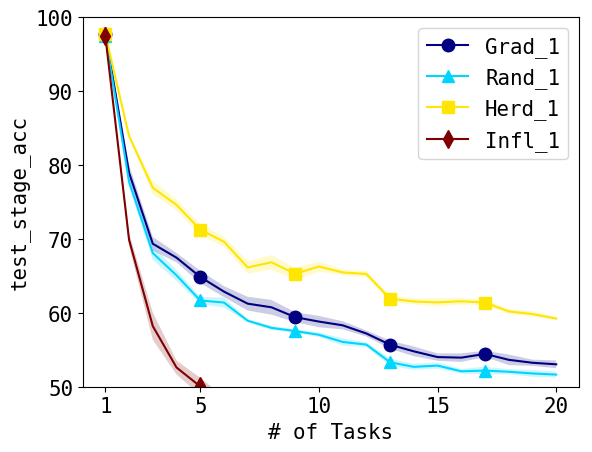}
    \caption{Task Average (1 point/class)}
    \label{subfig:mem1_test_stage_acc}
  \end{subfigure}
  \hfill
  \begin{subfigure}[b]{.65\columnwidth}
  \includegraphics[width=\textwidth]{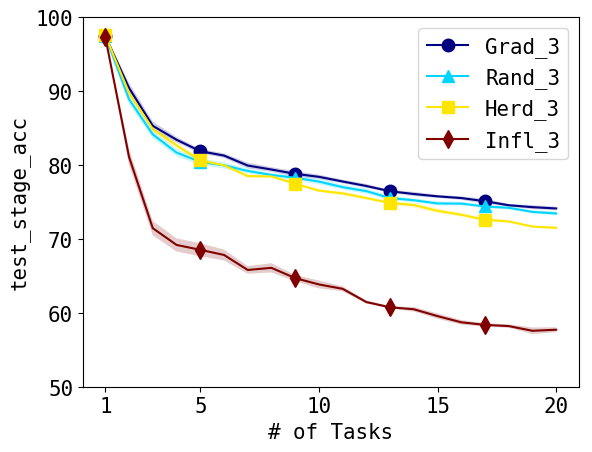}
    \caption{Task Average (3 points/class)}
    \label{subfig:mem3_test_stage_acc}
  \end{subfigure}
  \hfill
  \begin{subfigure}[b]{.65\columnwidth}
  \includegraphics[width=\textwidth]{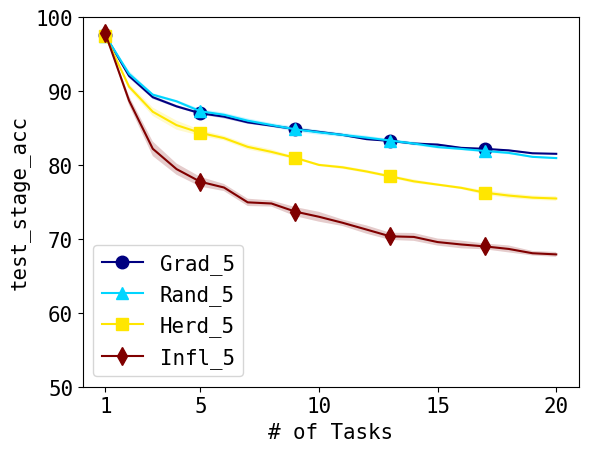}
    \caption{Task Average (5 points/class)}
    \label{subfig:mem5_test_stage_acc}
  \end{subfigure}
  \caption{Nonlinear MNIST Test Accuracy (Mean \& standard error over 5 runs)}
  \label{fig:select_points}
\end{figure}

We run the aforementioned methods on the Nonlinearly Transformed MNIST task and the results are shown in \cref{fig:select_points}.
First, note that when we only use 1 single point per class (a total memory of 10 points per task), herding outperforms other methods.
Essentially, with linear kernel, herding chooses the image that is closest to the mean of the class images.
However, its performs worse than Rand as we select more data points. 
Second, our proposed gradient-based method can outperforms the class-stratified random sampling when the memory is small,
but when the memory is sufficiently large, Rand and Grad become indistinguishable.
Finally, using influence function is the least favourable for all these cases.

Therefore, at the end of the day, class-stratified random sampling is still a simple yet effective way to select points into memory.

\section{Additional Results for Incremental \mbox{CIFAR100}}
\label{app:supp_cifar100}
Here we present additional results on the incremental CIFAR100 setting.

\begin{figure}[h]
\centering
  \includegraphics[width=.65\columnwidth]{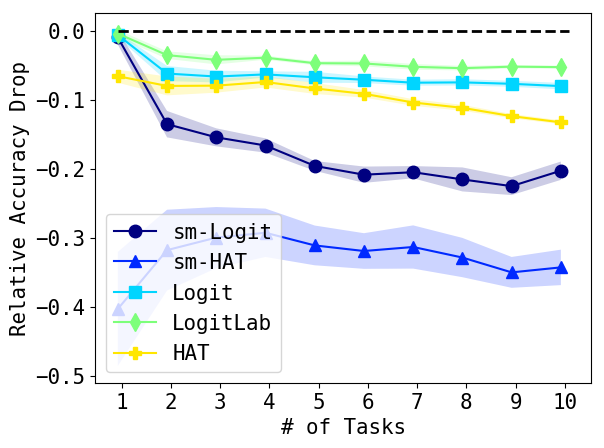}
  \caption{Average Forgetting Ratio for CIFAR100}
  \label{fig:supp_cifar100_rho}
\end{figure}

\textbf{Add True Labels}.
In addition to memorizing the logits, we can at the same time record the true labels of memory data.
Correspondingly, we add the KL divergence of memory data to the objective~(\ref{eq:matchinglogits}).
As shown in \cref{fig:supp_cifar100_rho}, LogitLab (Logit with additional true label) can noticeable outperform Logit alone. 

\textbf{Small Models}.
To see how the performance gap between FSR and HAT is accentuated when the model capacity is limited, we show the results with a smaller model in \cref{fig:supp_cifar100_rho} (with prefix ``sm'').
As we mentioned in the main text, the attention/gating mechanism of HAT can prevent effective representation sharing between tasks. 
When the model capacity is limited, HAT's performance can be seriously hampered.


\end{document}